\documentclass[letterpaper, 10 pt, conference]{ieeeconf}
    \IEEEoverridecommandlockouts
    \usepackage{times}
    \usepackage{amsmath}
    \usepackage{amssymb}
    \usepackage{graphicx}
    \usepackage{xcolor}
    \usepackage{algorithm}
    \usepackage[noend]{algpseudocode}
    \usepackage{balance}
    \usepackage{cuted}
    \usepackage{capt-of}
    \usepackage{booktabs}
    \usepackage{siunitx}
    \usepackage{multirow}
    \usepackage{booktabs}
    \usepackage{cite}
    \usepackage{tabularx}
    \usepackage{caption}
    \usepackage{url}

    \graphicspath{{figures/}}

    \def\figref#1{Fig.~\ref{#1}}
    \def\tabref#1{Tab.~\ref{#1}}
    \def\eqref#1{Eq.~(\ref{#1})}

    \newcommand\etal{\emph{et al.}}
    \DeclareMathOperator*{\argmax}{arg\,max}

    \def\argmax{\mathop{\rm argmax}}

    \definecolor{ostargreen}{HTML}{2ECC71}
    \definecolor{auxpurple}{HTML}{9B59B6}
    \definecolor{transred}{HTML}{E74C3C}
    \definecolor{llmblue}{HTML}{3498DB}
    \definecolor{uniformgray}{HTML}{7F8C8D}

        \title{\LARGE \bf O-STaR: Open-Vocabulary Object Search through \\
        Spatio-Temporal Reasoning on Furniture-Anchored Scene Graphs}
    \author{
  Rohit Menon \quad Gokul Krishna Chenchani \quad
  Yasmin Schmiede \quad Niklas Mueller-Goldingen \\[3pt]
  Hermann Blum$^*$ \quad Maren Bennewitz$^*$
}

\begin{document}
    \maketitle
    \thispagestyle{empty}
    \pagestyle{empty}
    \def\thefootnote{}\footnotetext{$^*$Equal Supervision. All authors are with the University of Bonn, Germany.
    R. Menon, M. Bennewitz and H. Blum are further affiliated with the Lamarr Institute and the Center for Robotics, Bonn.
    R. Menon, N. Mueller-Goldingen  and M. Bennewitz are also with the Humanoid Robots Lab.
    H. Blum is also with the Robot Perception and Learning Lab.\\
    This work was partially funded by the Deutsche Forschungsgemeinschaft (DFG, German Research Foundation) under Germany’s Excellence Strategy, EXC-2070 -- 390732324 -- PhenoRob, under grant 459376902 – AID4Crops, and under grant 571413906, and by the BMBF within the Robotics Institute Germany, grant No. 16ME0999.
}
    \renewcommand{\thefootnote}{\arabic{footnote}}

    \begin{abstract}
        Locating everyday objects in dynamic home environments is challenging for service robots because observations are sparse and objects may be relocated between searches, without the robot's knowledge.
        In this paper, we study on-demand episodic object search in semi-structured homes.
        To solve this task, we propose Open-vocabulary Object Search through Spatio-Temporal Reasoning~(\mbox{O-STaR}), which combines semantic placement priors, geometric pruning of compartments, and persistent object-location beliefs using sparse furniture-anchored scene graphs.
        A large language model (LLM) annotates the scene graph with inspection affordances, ranks furniture-level object locations, and provides an object-category size prior. O-STaR then grounds these outputs in the annotated scene graph and down-weights concealed locations that geometrically cannot contain the target.
        Target and incidental observations update the object-location beliefs across queries, while the affordances determine how locations are inspected.
        Long-term evaluations on HOMER+~\cite{patel2023predicting} and DynamicTHOR~\cite{wang2026dynamicthor} show that O-STaR significantly outperforms static and push-back LLM ranking methods.
        Furthermore, real-world robotic experiments demonstrate that our framework supports scene-graph-based search reasoning and embodied inspection, increasing search success from 53.3\% for a baseline inspection policy to 86.7\% for O-STaR.
    \end{abstract}

    \section{Introduction}
    \label{sec:intro}

    Recent advances in autonomous systems are bringing robots closer to deployment in household environments.
    These environments are highly dynamic~\cite{miller2018understanding}, as objects are frequently moved and relocated, while many of these movements are not directly observed by the robot.
    Therefore, for assistive behaviors such as fetching, tidying, and helping to recover misplaced items~\mbox{\cite{hochschild2012second}}, reliable object search is a prerequisite for robust robot behavior.
    However, existing work typically assumes either fully static environments~\cite{gu2024conceptgraphs, liu2024okrobot} or completely unknown spaces that require exhaustive exploration strategies~\cite{zheng2023system}.
    In practice, households are semi-structured, i.e., rooms and furniture remain largely static, while objects are routinely relocated or placed within furniture, for example inside cabinets, drawers, or other enclosed spaces.
    \begin{figure}[t]
      \centering
      \includegraphics[width=0.97\columnwidth, angle=0]{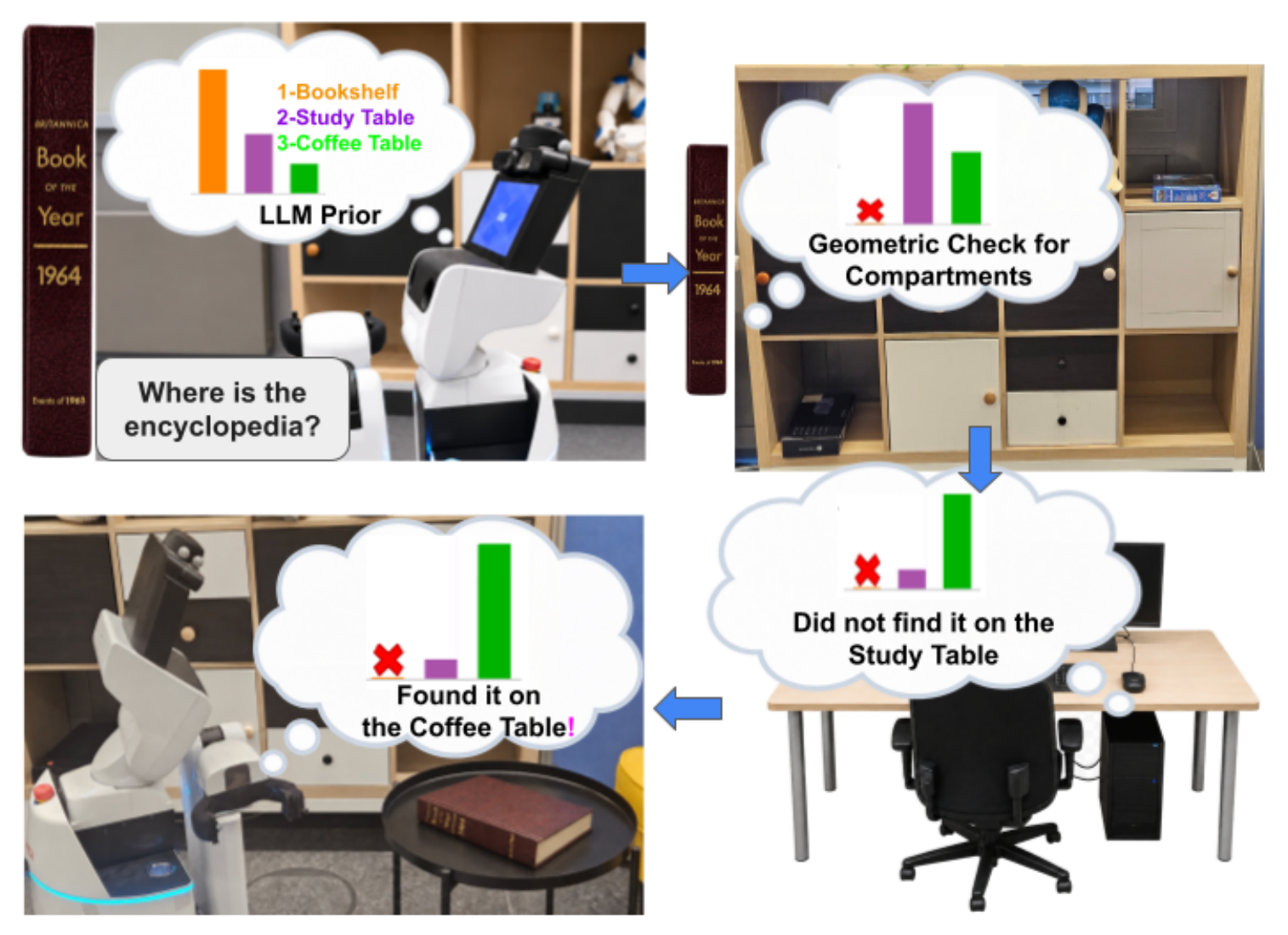}
      \caption{\small {Illustrative O-STaR search for an encyclopedia.
        The initial semantic prior ranks the bookshelf, study table, and coffee table as plausible locations, but geometric pruning excludes the bookshelf's compartments since they cannot accommodate the encyclopedia.
        The robot therefore inspects the study table first. After a negative observation, the belief is updated and the most likely location is now the coffee table, where the encyclopedia is subsequently found.}
        }
      \label{fig:motivation}
     \vspace{-16px}
    \end{figure}
    Consequently, service robots deployed in dynamic home environments must operate on \emph{outdated} world beliefs during object search and prioritize likely locations, as exhaustive inspection is prohibitively time-intensive.

    To this end, recent approaches use Large Language Models~(LLMs) for common-sense object-placement priors~\mbox{\cite{qiu2024jsr1,honerkamp2024momallm}}, but capture only population-level behavior and no household-specific patterns or environment-specific physical feasibility.
    Furthermore, to handle relocations outside the robot's field of view, existing works perform costly updates of dense semantic maps~\cite{liu2024dynamem,qiu2024jsr1} or active re-exploration strategies~\cite{bogenberger2026glasses}, requiring autonomous sensing beyond user-initiated search and conflicting with demand-driven household operation.

    In this paper, we propose Open-vocabulary Object Search through Spatio-Temporal Reasoning~(\mbox{{O-STaR}}), which represents the environment as a fixed 3D semantic scene graph~\cite{gu2024conceptgraphs} of rooms, furniture, and searchable locations.
    \mbox{O-STaR} maintains Dirichlet--categorical location beliefs only for queried object categories, avoiding exhaustive open-world object discovery and tracking.
    Inspection provides \mbox{hit-or-miss} evidence for the target object and simultaneously updates beliefs of previously queried categories.
    Our framework combines semantic priors, geometric feasibility checks, and persistent belief updates to adapt to household-specific object locations.
    Searchable locations encode inspection affordances for exposed surfaces, open interiors, and concealed compartments, with the latter additionally specifying manipulation actions.
    Thus, the scene graph represents static environment structure, while object-location beliefs capture its dynamic state.

    In summary, this work makes the following contributions:
    \begin{enumerate}
    \item A hierarchical furniture-anchored scene graph of exposed surfaces, open interiors, and concealed searchable locations with inspection and manipulation affordances.
    \item A Dirichlet--categorical belief model initialized from language-model priors and persistently updated from target and incidental observations to learn household-specific object-location patterns.
    \item Object-conditioned geometric pruning of concealed compartments and affordance-conditioned inspection of searchable locations.
    \item A robot-agnostic object-search framework operating directly on a sparse scene graph without a dense 3D semantic map, demonstrated on the Toyota HSR~\cite{yamamoto2018human}.
    \end{enumerate}

    Through matched evaluations over 60-day sequences in three HOMER+~\cite{patel2023predicting} households and five-day sequences in all 100 DynamicTHOR~\cite{wang2026dynamicthor} scenes, we show that semantic initialization improves early search success, persistent belief updates improve upon static and push-back language-model rankings, and incidental observations provide further gains under perfect and noisy observations.

    \section{Related Work}
    \label{sec:related}
    \subsection{Classical Object Search Approaches}
    Robotic object search has long been formulated as active perception under partial observability, with semantic object--location relations guiding sensing and belief updates~\mbox{~\cite{aydemir2011search,kunze2014using,hanheide2011probabilistic,wandzel2019multi,zeng2020semantic}}.
    These formulations typically revise candidate-location beliefs within a search episode.
    Long-term autonomy instead models recurring place or activity patterns across deployments~\cite{hawes2017strands,krajnik2017fremen}, rather than O-STaR's persistent query-specific beliefs over furniture and compartment locations.

    \subsection{Open Vocabulary Object Search}
    Foundation models enable open-vocabulary object search by combining semantic priors with structured 3D representations~\cite{gu2024conceptgraphs,liu2024okrobot}.
    ConceptGraphs~\cite{gu2024conceptgraphs}, HOV-SG~\cite{werby23hovsg}, and Open Scene Graphs~\cite{loo2025openscenegraphs} provide object- and hierarchy-level scene representations for semantic reasoning and object-goal navigation, while osmAG-LLM~\cite{xie2026osmag} uses LLM reasoning over semantic maps for displaced or unmapped targets.
    JSR-1~\cite{qiu2024jsr1}, MoMa-LLM~\cite{honerkamp2024momallm}, and SCOUT~\cite{mahdi2026relational} extend such representations to interactive search, with SCOUT using distilled relational semantics and coarse node-level affordances to guide exploration.
    In contrast, O-STaR initializes query-specific furniture-level priors and maintains persistent object-location beliefs that adapt from search observations across relocations, while explicitly modeling compartment-level manipulation and inspection affordances.
    It additionally prunes searchable locations that cannot geometrically accommodate the target before inspection, using LLM-estimated object category size priors.

    Dynamic mapping approaches maintain object-level world models as the environment changes.
    DynaMem updates a 3D spatio-semantic memory from observed object changes~\cite{liu2024dynamem}, while \emph{Where Did I Leave My Glasses?}~\cite{bogenberger2026glasses} additionally models the stationarity of out-of-view objects and actively revisits stale regions.
    O-STaR instead keeps the structural scene graph fixed and represents unobserved object relocation through persistent location beliefs updated only from episodic search observations.

    \subsection{Household Adaptive Search}
    HOMER~\cite{patel2022proactive} and HOMER+~\cite{patel2023predicting} condition spatio-temporal predictions on user actions and queries, potentially requiring dedicated and intrusive sensing of human activities.
    PerSim~\cite{li2026personalize} learns personality-conditioned room and co-occurrence priors, while Chikhalikar~\etal~\cite{chikhalikar2025enhancing} personalize object--landmark--room relations.
    In contrast, O-STaR learns household-specific furniture- and compartment-level location beliefs initialized from LLM priors solely through observations acquired during object searches, without requiring continuous monitoring of human activities or object relocations.

    \subsection{Active and Interactive Perception}
    Active and interactive perception mitigate occlusion through informative viewpoints or environment manipulation~\cite{zheng2021multi,koc2024objectaware,dengler2025efficient}.
    HIMOS~\cite{schmalstieg2023learning} and Spot-Compose~\cite{lemke2023spotcompose} integrate object search with articulated-object manipulation.
    Similarly, Stretch-Compose~\cite{menon2025open} uses fixed LLM-based ranking of furniture locations for dynamic and concealed-space search, however without any persistent belief adaptation.
    O-STaR instead selects furniture or compartments from a persistent object-location belief, while inspection affordances determine whether sensing is sufficient or requires accessing a concealed location.

    To the best of our knowledge, O-STaR is the first long-term object-search framework to perform belief-conditioned, affordance-specific active inspection directly over a sparse 3D furniture scene graph, using only a 2D occupancy map for navigation and without a dense semantic or voxel map.

    \section{Problem Statement}
    \label{sec:problem}

    We consider open-vocabulary search for an object category~$o$ in a furniture-anchored scene graph $G=(\mathcal{N},\mathcal{E})$, constructed during offline environment pre-processing.
    The node set $\mathcal{N}$ represents rooms, furniture anchors, and their compartments, while the edge set $\mathcal{E}$ encodes their hierarchical relations.
    The finite set of searchable regions $\mathcal{L}=\{l_1,\ldots,l_L\}$ comprises exposed support surfaces, open interiors, and concealed compartments.
    We assume a singleton target, such that exactly one instance of category~$o$ occupies one region $l^\star\in\mathcal{L}$ during each search episode.
    Objects get relocated between episodes without being observed, while the locations of rooms, furniture, and compartment geometry remain fixed.
    Each region $l_i$ has an inspection affordance $\rho_i$ and, if concealed, additionally a manipulation affordance $\mu_i$.
    An inspection action produces a noisy observation indicating whether the queried object is detected at the inspected region.
    The robot's objective is to identify $l^\star$ while minimizing navigation, inspection, and manipulation cost during search, under a fixed search budget.

    \section{Our Approach}
    \label{sec:approach}
    O-STaR represents a household as a static furniture-anchored scene graph and maintains a persistent belief, using Dirichlet pseudo-counts over searchable locations for each queried object category.
    The scene graph compactly captures the relatively static room--furniture structure and provides a structured interface for LLM-based annotation and querying.
    The LLM assigns inspection affordances and, for each queried category, provides the top-three furniture ranking and size and orientation priors.
    For concealed locations, we infer manipulation affordances. 
    The furniture ranking and geometric feasibility constraints initialize the object-location belief, which is updated from inspection outcomes across search episodes.
    During search, the robot selects the next location to inspect based on the belief, and the inspection and manipulation affordances determine how to observe and access the search locations.
    The following sections detail these components.

    \subsection{Affordance Annotated Furniture-Anchored Scene Graph}
    \label{sec:preprocess}
    \begin{figure}[t]
        \centering
        \includegraphics[width=0.97\columnwidth]{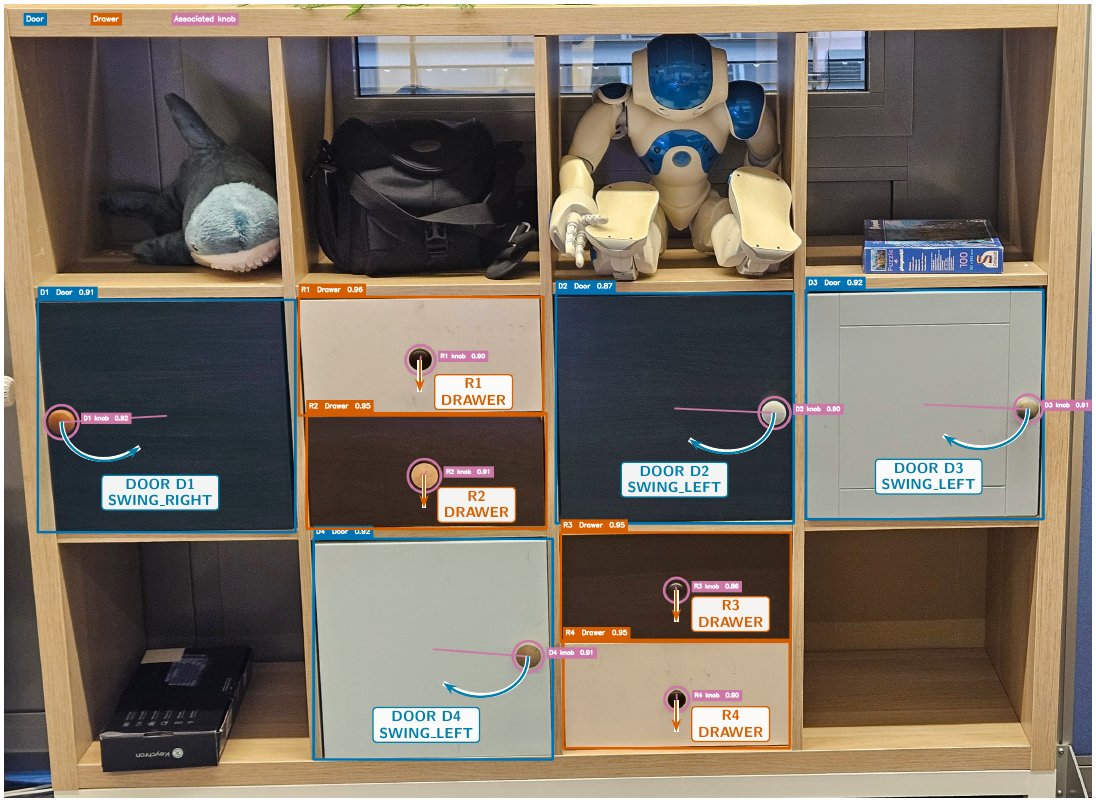}
        \caption{\small {Detection of concealed compartments and inference of their manipulation affordances from associated knob locations.
        Drawers are identified by approximately centered knobs, whereas laterally positioned knobs indicate hinged doors.
        The knob position determines the opening direction: D1 is inferred to swing right, while D2--D4 are inferred to swing left.}}
        \label{fig:bookshelf_doors_drawers}
        \vspace{-18px}
    \end{figure}
\begin{figure}[t]
        \centering
        \includegraphics[width=0.97\columnwidth]{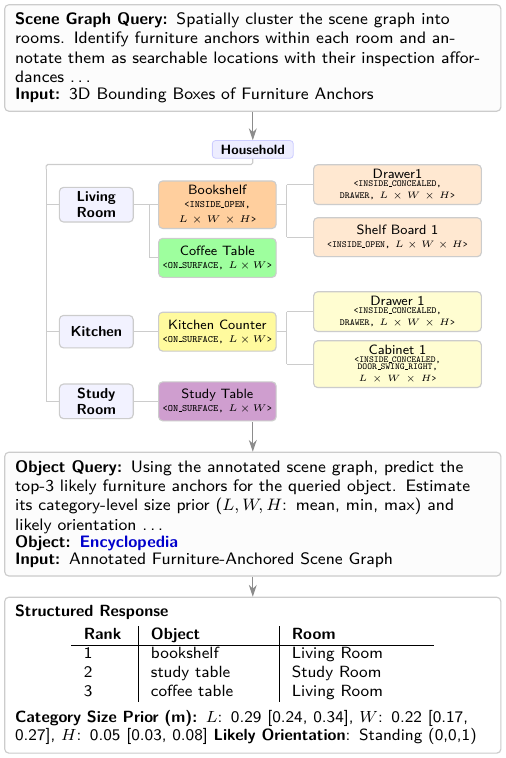}
        \caption{\small{Scene-graph and object-query pipeline.
Furniture anchors are organized by room and annotated with searchable locations, inspection/manipulation affordances, and geometric extents; for a target object, the annotated graph is queried to rank the top-3 likely furniture anchors and estimate a category-level size prior for subsequent geometric reasoning.}}
        \label{fig:llm_query}
        \vspace{-18px}
    \end{figure}

    We assume a furniture-anchored scene graph $G=(\mathcal{N},\mathcal{E})$ obtained from an aligned 3D scan, with furniture instances extracted, e.g., using open-vocabulary segmentation~\mbox{\cite{schult2023mask3d,rozenberszki2022language}}.
    Each room, furniture, and compartment node has a semantic label, centroid $\bar{\mathbf{x}}_i\in\mathbb{R}^3$, and 3D bounding box $\mathbf{B}_i$, while edges encode their hierarchical relations.
    Let $\mathcal{F}\subseteq\mathcal{N}$ denote the furniture anchors and $\mathcal{L}\subseteq\mathcal{N}$ the searchable locations.
    For each $f\in\mathcal{F}$, $\mathcal{C}(f)\subseteq\mathcal{L}$ comprises the searchable anchor, if applicable, and its independently searchable leaf compartments.
    Anchor-level locations represent exposed surfaces or open interiors, whereas independently accessible concealed compartments are represented as separate leaf locations.
    For example, a kitchen counter may comprise an anchor-level work surface together with leaf locations for individual drawers and door-enclosed compartments. To specify how a location can be inspected, each $l_i\in\mathcal{L}$ is assigned the inspection affordance~$\rho_i$~(see \figref{fig:llm_query}) with

    {\small
    \begin{equation*}
    \vspace{-3px}
    \rho_i \in \{ \texttt{ON\_SURFACE}, \texttt{INSIDE\_OPEN}, \texttt{INSIDE\_CONCEALED}\}
    \end{equation*}
    }
    For concealed locations, we apply SAM3~\cite{carion2025sam3} to segment doors and drawers together with their knobs or handles, as shown in~\figref{fig:bookshelf_doors_drawers}.
    The relative handle position determines the manipulation affordance

    {\small
    \begin{equation*}
    \vspace{-3px}
    \mu_i\in\{\texttt{DOOR\_SWING\_LEFT},\ \texttt{DOOR\_SWING\_RIGHT},\ \texttt{DRAWER}\},
    \vspace{-4px}
    \end{equation*}}
    which specifies the access action and inspection direction.
    Doors are inspected from the front and drawers from above.
    Thus, these annotations are inferred directly from visual observations without artificial fiducial markers, unlike SCOUT~\cite{mahdi2026relational} and MoMaLLM~\cite{honerkamp2024momallm}.

    \subsection{Object-Conditioned Belief Initialization}
    \label{sec:belief_initialization}

    \paragraph{Furniture-Level Semantic Prior for Placement}
    For a target object category $o$, O-STaR asks an LLM to rank the three furniture anchors with the highest placement plausibility from the valid room--furniture context as $f_1 \succ f_2 \succ f_3$.
    We retain only the top three to capture the strongest semantic associations while limiting dependence on increasingly uncertain lower-ranked predictions.
    Each ranked furniture anchor is subsequently mapped to its searchable locations in~$\mathcal{C}(f)$.
    We query at the furniture level because common-sense placement relations apply to furniture categories rather than individual compartment indices.
    Since LLM scores are not calibrated against household-specific placement frequencies, we retain only the ranking and assign exponentially decaying weights
    {\small
\begin{equation}
    \label{eq:decay}
    w_k=e^{-\tau(k-1)}, \qquad k\in\{1,2,3\}, \qquad \tau>0
    \end{equation}}
    where $\tau$ controls the decay between successive ranks.
    We normalize the furniture weights as $\bar{w}_k=w_k/\sum_{r=1}^{3}w_r$.
    These weights allocate the total semantic prior mass $S_0$ among the ranked furniture anchors after geometric feasibility checks and renormalization.

    \paragraph{Orientation-Aware Geometric Feasibility Check}
    The semantic ranking identifies plausible furniture anchors but does not determine whether an object can be plausibly stored inside a concealed compartment.
    O-STaR therefore applies geometric feasibility checks only to concealed compartments, while retaining open shelves and exposed surfaces. The LLM provides category-level minimum dimensions $\mathbf{d}_{\min}^o$, mean dimensions $\mathbf{d}_{\mathrm{mean}}^o$, and a typical storage orientation when known.
    A compartment is rated \emph{zero} if $\mathbf{d}_{\min}^o$ does not fit under any axis permutation, \emph{low} if $\mathbf{d}_{\mathrm{mean}}^o$ does not fit or fits only outside the typical orientation, and \emph{high} if $\mathbf{d}_{\mathrm{mean}}^o$ fits in the typical orientation.
    If orientation is unknown, fitting the mean dimensions under any permutation yields a high rating.
    Zero-rated compartments are removed, while low-rated compartments retain a fraction $\eta\in(0,1)$ of their semantic mass.
    Let
    {\small
\begin{equation}
    \label{eq:semantic_mass}
     s_{0,i}^{o} = \frac{S_0\bar{w}_k}{|\mathcal{C}(f_k)|}
    \end{equation}}
    denote the semantic mass initially assigned to location $l_i\in\mathcal{C}(f_k)$.
    With $\lambda_i^o\in\{0,\eta,1\}$ for zero-, low-, and high-rated concealed locations, respectively, and $\lambda_i^o=1$ elsewhere, the mass removed from anchor $f_k$ is
    {\small
\begin{equation}
    \label{eq:mass_removal}
     \Delta_k^o = \sum_{i:\, l_i\in\mathcal{C}(f_k)} (1-\lambda_i^o)s_{0,i}^{o} \end{equation}}
    This mass is redistributed among its high-rated locations $\mathcal{H}_k^o$, yielding
    {\small
\begin{equation}
    \label{eq:mass_redistr}
    \alpha_{0,i}^{o}
    =
    1 
    +
    \begin{cases}
        s_{0,i}^{o} + \dfrac{\Delta_k^o}{|\mathcal{H}_k^o|},
        & l_i \in \mathcal{H}_k^o, \\[7pt]
        \lambda_i^o s_{0,i}^{o},
        & \text{otherwise},
    \end{cases}
    \end{equation}}
     Thus, the geometric check modifies the location-level prior while preserving the semantic mass and ranking of each furniture anchor.
     Feasible locations outside the ranked furniture anchors are initialized with $\alpha_{0,i}^o=1$.

    \subsection{Persistent Belief Updates}
    \label{sec:phase2}

    Under the singleton assumption, O-STaR models the location of each initialized object category $o$ using a Dirichlet--categorical model over its feasible search set $\mathcal{L}^o$.
    The Dirichlet pseudo-counts $\boldsymbol{\alpha}_t^o$ persist across search episodes $t$ and accumulate household-specific evidence from individual inspections.
    The corresponding categorical belief is
    {\small
\begin{equation}
    b_t^o(l_i)
    =
    \frac{\alpha_{t,i}^o}
    {\sum_{j:\,l_j\in\mathcal{L}^o}\alpha_{t,j}^o}.
    \end{equation}}
    Let $\mathcal{I}_t^o$ denote the set of locations already inspected for category $o$ during the current episode, and let $\mathcal{U}_t^o = \mathcal{L}^o \setminus \mathcal{I}_t^o$ denote the remaining uninspected feasible candidate locations.
    A hit reinforces the inspected location, whereas a miss distributes total evidence $w_{\mathrm{miss}}$ uniformly over the remaining alternatives in $\mathcal{U}_t^o$, reducing the inspected location's normalized belief without subtracting pseudo-counts.
   {\small
\begin{equation}
    \alpha_{t+1,j}^{o}
    =
    \alpha_{t,j}^{o}
    +
    \begin{cases}
    w_{\mathrm{hit}},
    & z_{t}^{o}(l_i)=1,\ j=i,\\
    \dfrac{w_{\mathrm{miss}}}{|\mathcal{U}_{t}^{o}|},
    & z_{t}^{o}(l_i)=0,\ l_j\in\mathcal{U}_{t}^{o},\\
    0,
    & \text{otherwise}
    \end{cases}
    \label{eq:observation_update}
    \end{equation}}
    for binary inspection outcomes $z_t^o(l_i)$.
    The inspected set is reset between episodes, while $\boldsymbol{\alpha}_t^o$ persists. We also evaluated an inter-episode temporal decay factor to discount stale observations over time ($\boldsymbol{\alpha}_{t+1}^o = (1-\gamma)\boldsymbol{\alpha}_t^o + \gamma\mathbf{1}$).
    However, in our demand-driven search regime, observations are inherently sparse and we found that decaying pseudo-counts did not produce any significant advantage for long-term belief update.
    Consequently, O-STaR maintains persistent accumulation across episodes.

    \subsection{Incidental Observations}
    \label{sec:aux_obs}
    An inspection for the current target can also provide evidence about previously queried object categories within the same searchable location.
    O-STaR uses this evidence to update persistent object--location beliefs without additional navigation or manipulation.

    Let $\mathcal{M}_t$ denote the object categories queried in episodes preceding episode $t$, excluding the current target category $o$.
    After inspecting location $l_i$, the acquired observation is evaluated for the current target $o$ and each \mbox{$o'\in\mathcal{M}_t$} for which $l_i$ is feasible. 
    A detection provides incidental hit evidence, while a non-detection provides miss evidence. 
    Each valid incidental observation updates $\boldsymbol{\alpha}_{t}^{o'}$ according to~\eqref{eq:observation_update} using the incidental evidence weights.
    Thus, one inspection can update the persistent beliefs of multiple previously queried categories.

   \subsection{Affordance-Conditioned Inspection Action Selection}
    \label{sec:active_inspection}
    \begin{table}[t]
    \centering
    \small
    \setlength{\tabcolsep}{2.5pt}
    \setlength{\aboverulesep}{0pt}
    \setlength{\belowrulesep}{0pt}
    \renewcommand{\arraystretch}{1.3}

    \begin{tabular}{
        m{0.23\columnwidth}|
        m{0.36\columnwidth}|
        m{0.36\columnwidth}
    }
        \toprule
        Affordance & Inspection region & Observation action \\
        \midrule
        \texttt{ON\_SURFACE} &
        volume above the \mbox{support} top &
        nominal, raised, or alternate-side view \\
        \midrule
        \texttt{INSIDE\_OPEN} &
        annotated open interior &
        full-face and, when \mbox{required}, tiled views \\
        \midrule
        \shortstack[l]{\texttt{INSIDE\_}\\\texttt{CONCEALED}} &
        compartment interior &
        open using $\mu_i$ and \mbox{inspect} \\
        \bottomrule
    \end{tabular}
    \caption{Affordance-conditioned inspection actions $\rho_i$.}
    \label{tab:affordance_actions}
    \vspace{-18px}
\end{table}

\mbox{O-STaR} prioritizes the inspection of exposed locations over concealed compartments due to the additional execution cost of manipulation.
Let $\mathcal{F}_t^o$ denote the set of remaining anchors available for open inspection and $\mathcal{A}_t^o(f)$ the set of remaining searchable concealed compartments of anchor $f$.
The next anchor available for open inspection is selected as
\begin{equation}
  f_t^{\mathrm{next}} =
  \argmax_{f\in\mathcal{F}_t^o} b_t^o(f).
  \label{eq:anchor_selection}
\end{equation}
Each anchor is initially inspected according to its inspection affordance, e.g., full-volume inspection for \texttt{ON\_SURFACE} and full-face inspection for \texttt{INSIDE\_OPEN}.
Following this inspection, if the current anchor $f_t$ contains remaining searchable concealed compartments, we select
\begin{equation}
  l_t =
  \argmax_{l_i\in\mathcal{A}_t^o(f_t)} b_t^o(l_i).
  \label{eq:compartment_selection}
\end{equation}
The selected compartment is opened and inspected only if
$b_t^o(l_t)>2b_t^o(f_t^{\mathrm{next}})$.
This criterion is reevaluated after each compartment inspection until no remaining compartment satisfies it, at which point the robot proceeds to the next anchor.

The inspection affordance $\rho(l)$ of the selected location $l$, whether an anchor or a compartment, determines the observation volume and required action as specified in Table~I.
Candidate inspection poses are generated considering reachability and the robot's embodiment.
Broadside views are preferred for exposed surfaces, whereas open interiors are viewed along their shallow open faces.
Candidate viewpoints are scored based on geometric obstruction and non-target clutter within the observation volume.
To improve visibility while limiting execution cost, viewpoint refinement is enabled only if
\begin{equation}
  b_t^o(l) \geq \frac{\beta}{N^o}
  \label{eq:viewpoint_refinement}
\end{equation}
where $N^o$ is the number of searchable locations and $\beta$ is a scaling factor relative to the uniform belief.
Refinement prioritizes elevated or tilted viewpoints from the same side before considering alternative sides.
Target and incidental inspection outcomes update the beliefs as described in Secs.~IV-C and IV-D.
    \section{Experimental Evaluation}
    \label{sec:experiments}

    We evaluate two complementary aspects of O-STaR.
    First, we evaluate long-term object-location reasoning over repeated searches and relocations, including semantic initialization, persistent belief adaptation, and incidental observations using two datasets, namely HOMER+~\cite{patel2022proactive,patel2023predicting} and DynamicTHOR~\cite{wang2026dynamicthor}.
    Second, we perform real-world experiments with an HSR platform to evaluate robotic inspection of exposed and concealed locations, geometric feasibility checks,  and inspection policies.

    We investigate the following research questions.
    \textbf{RQ1:} Does language-model initialization improve search compared to a uniform prior, particularly during early repeated searches?
    \textbf{RQ2:} Do persistent target-only belief updates improve over fixed and push-back LLM rankings as searches accumulate?
    \textbf{RQ3:} Do incidental observations provide additional long-term benefit beyond target-only updates?
    \textbf{RQ4:} Do LLM-derived size priors preserve object--compartment fit decisions and enable conservative geometric pruning without erroneously removing feasible search locations?
    \textbf{RQ5:} Can sparse scene graphs with affordance annotations support real-world robotic object search, and do affordance-conditioned inspection policies improve search success over naive furniture-centered views?
    RQ1, RQ2, RQ3 are evaluated in simulation whereas RQ4 and RQ5 are validated in the real world.

    \subsection{Long-Term Observation Data}
    \label{sec:exp_setup}

    The HOMER+~\cite{patel2023predicting} dataset contains three households with repeated object--furniture relocations.
    We replay 60 days per household on a fixed furniture scene graph, using 27 common objects, ten queries per day, and three randomized query schedules (seeds 42--44), resulting in 600 search episodes per seed and noise setting.
    DynamicTHOR~\cite{wang2026dynamicthor} provides 50 normal and 50 hard ProcTHOR scenes; normal scenes contain one or two residents, whereas hard scenes contain three or four, inducing more frequent and complex relocations.
    For each scene, we replay five days, select six targets, and issue 20 object search queries per schedule, yielding 6,000 episodes per method and regime, split equally across difficulty subsets.
    We evaluate clean observations, 5\% false negatives (FN5), and 5\% false negatives with 5\% false positives (FN5+FP5).
    Inspection outcomes $z_t^o(l_i)$ are generated from the ground-truth placements with errors sampled according to the noise setting, while object-location beliefs persist throughout each sequence.

    \subsection{Baselines and Ablations}
    \label{sec:baselines}
    O-STaR uses the DeepSeek-V4 Flash~\cite{xu2026deepseek} API to generate a furniture-level semantic ranking once per target category and scene graph.
    All rankings are generated and cached before evaluation so that matched methods use identical semantic priors.
    We compare O-STaR against: \textbf{Uniform}, which uses a uniform prior with the same persistent belief updates and incidental observations as O-STaR; \textbf{Static LLM}, which just follows the fixed LLM ranking; and \textbf{Push-back LLM}, which moves an inspected miss to the end of the ranking.
    We additionally evaluate \textbf{O-STaR w/o IO}, which retains only target observations and omits incidental observations of previously queried objects.
    All methods inspect the highest-scoring affordable~(cf. Eq.~\ref{eq:budget}) unvisited location under the same budget $B=600\,\mathrm{s}$.
    \subsection{Experimental Parameters}
    For both benchmarks and all noise settings, the semantic prior mass is $S_0=300$, decay factor $\tau = 1$ in \eqref{eq:decay}, $\beta = 1.5$ in \eqref{eq:viewpoint_refinement}, and the geometric feasibility down-weighting factor $\eta=0.1$ in \eqref{eq:mass_removal}, \eqref{eq:mass_redistr}.
    Target observations use $(w_{\mathrm{hit}},w_{\mathrm{miss}})=(100,200)$, whereas incidental observations use $(25,50)$.
    The parameters were determined heuristically using only a subset of HOMER+'s Household B graphs.
    We calculate the search cost as follows:
    {\small
\begin{equation}
    c_t(l_i)=c_{\mathrm{nav}}\!\left(d_t(l_i)\right)
    +c_{\mathrm{manip}}(l_i)+c_{\mathrm{insp}}.
    \label{eq:experimental_action_cost}
    \end{equation}}
    where $d_t(l_i)$ is the A* path length from the current robot pose to the inspection pose.
    For $d\geq0.01 \mathrm{m}$, $c_{\mathrm{nav}}(d)=d/0.05+5$\,s.
    The inspection cost is $c_{\mathrm{insp}}=2.5$\,s.
    Manipulation costs are 60\,s for drawers, 90\,s for doors, and zero for exposed surfaces and open interiors.
    A location is affordable if
    {\small
\begin{equation}
    c_t(l_i)
    \leq
    B-\sum_{r<t}c_r(l_r).
    \label{eq:budget}
    \end{equation}}

    \subsection{Metrics}
    \label{sec:exp_metrics}
    We report search success, inspected locations per episode, and simulated search time, including failed episodes.
    Success is the fraction of target-present queries in which the target is found within the budget.
    Simulated time is the accumulated action cost from~\eqref{eq:experimental_action_cost}.
    We also performed paired Wilcoxon signed-rank tests across 50 scene-level pairs for DynamicTHOR and nine household--schedule pairs for HOMER+, with Holm correction applied to account for multiple comparisons.
    \subsection{Long-Term Belief Results}
    \label{sec:exp_results}
    \noindent\textbf{Language-model prior alignment:}
    We first quantify how well the fixed LLM ranking agrees with the ground-truth target location before any belief updates.
    For the 27 HOMER+ object categories common to all three households, Top-1/Top-3 alignment is 44.4\%/66.7\% in Household~A, 18.5\%/44.4\% in Household~B, and 29.6\%/37.0\% in Household~C.
    Across the 6,000 scheduled DynamicTHOR queries, Top-1/Top-3 alignment is 46.10\%/72.57\%, with Top-3 alignment of 76.23\% in normal and 68.90\% in hard scenes.
    These values characterize the quality of the zero-shot semantic prior before adaptation.
    \begin{table*}[t]
    \centering
    \scriptsize
    \setlength{\tabcolsep}{3pt}
    \begin{tabularx}{\textwidth}{@{}p{2.5cm}X*{3}{>{\centering\arraybackslash}p{3.05cm}}@{}}
    \toprule
    & & \multicolumn{3}{c}{Success (\%) / visits per episode / time per episode (s)} \\
    \cmidrule(lr){3-5}
    Dataset & Method & Clean & FN5 & FN5+FP5 \\
    \midrule
    \multirow{5}{=}{%
    \raggedright
    HOMER+~\cite{patel2023predicting}\\ All households\\ (A,B,C)}
    & Uniform Prior + belief + incidental & 91.74/1.36/238.4 & 87.19/1.50/256.0 & 85.20/1.47/258.6 \\
    & Static LLM ranking & 69.13/2.63/367.1 & 66.04/2.73/377.2 & 63.43/2.54/357.8 \\
    & Push-back LLM ranking & 76.26/1.98/302.0 & 72.52/2.12/315.9 & 71.81/1.99/304.8 \\
    & O-STaR w/o IO & 89.52/1.45/250.2 & 84.56/1.62/268.9 & 79.91/1.66/277.1 \\
    & \textbf{O-STaR (LLM prior+belief+incidental)} & \textbf{94.35/1.29/231.1} & \textbf{89.69/1.44/250.4} & \textbf{87.28/1.44/253.1} \\
    \midrule
    \multirow{5}{=}{%
    \raggedright{%
    DynamicTHOR~\cite{wang2026dynamicthor}\\ Normal (50)}
    }
    & Uniform Prior + belief + incidental & 62.47/4.45/309.1 & 58.23/4.76/330.4 & 50.27/4.46/333.3 \\
    & Static LLM ranking & 76.60/4.08/257.2 & 72.53/4.40/276.6 & 67.97/3.65/242.9 \\
    & Push-back LLM ranking & 78.03/3.50/219.9 & 73.47/3.90/244.9 & 70.00/3.34/223.1 \\
    & O-STaR w/o IO & 80.70/3.11/217.2 & 75.43/3.54/245.1 & 71.97/3.11/225.7 \\
    & \textbf{O-STaR (LLM prior+belief+incidental)} & \textbf{86.73/2.57/172.6} & \textbf{81.97/2.96/199.1} & \textbf{77.80/2.67/189.4} \\
    \midrule
    \multirow{5}{=}{%
    \raggedright{%
    DynamicTHOR~\cite{wang2026dynamicthor}\\ Hard (50)}
    }
    & Uniform Prior + belief + incidental & 39.50/5.18/423.8 & 37.97/5.27/432.3 & 29.70/4.83/440.7 \\
    & Static LLM ranking & 61.63/4.70/342.8 & 58.70/4.89/354.0 & 56.63/4.05/325.2 \\
    & Push-back LLM ranking & 63.20/4.27/319.3 & 59.83/4.50/334.9 & 58.17/3.75/311.2 \\
    & O-STaR w/o IO & 64.57/3.75/325.4 & 61.60/3.95/337.6 & 59.20/3.56/321.1 \\
    & \textbf{O-STaR (LLM prior+belief+incidental)} & \textbf{69.27/3.43/297.3} & \textbf{65.50/3.66/314.5} & \textbf{62.67/3.34/301.1} \\
    \bottomrule
    \end{tabularx}
    \caption{\small{Long-term object-search performance on HOMER+ and DynamicTHOR under clean, FN5, and FN5+FP5 observations.
    Each entry reports success rate (\%), mean inspected locations (visits), and mean simulated episode time (s); bold denotes the best result per condition.
    HOMER+ results aggregate 5,400 episodes, while each DynamicTHOR difficulty subset aggregates 3,000 episodes.
    On DynamicTHOR, paired Wilcoxon signed-rank tests across 50 scenes showed that O-STaR with IO significantly outperformed Static LLM, Push-back LLM, and O-STaR w/o IO i.e. without incidental observations in all three noise settings (Holm-adjusted \(p<0.05\)). On HOMER+, the same comparisons were significant across nine household--schedule runs (Holm-adjusted \(p<0.05\)), although these runs cover only three distinct households.}}
    \label{tab:homer_results}
    \vspace{-12px}
    \end{table*}

    \begin{figure*}[t]
        \centering
        \includegraphics[width=0.98\textwidth]{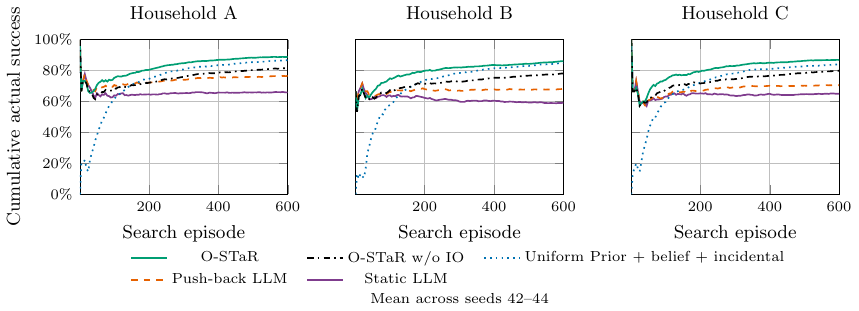}
        \caption{\small{Cumulative search success over the 600 search episodes for HOMER+ in each household under 5\% false negatives and 5\% false positives, with 600\,s search budget. Curves show means over seeds 42, 43, and 44. The LLM prior provides high initial success, while persistent belief updates improve success over the static and push-back rankings as observations accumulate. The uniform prior approaches the O-STaR methods only after a longer initial learning period.}}
        \label{fig:homer_cumulative_success}
         \vspace{-16px}
    \end{figure*}
    \noindent\textbf{RQ1 Language-model prior:}
    Semantic initialization primarily improves cold-start search and data efficiency rather than late-sequence performance.
    On HOMER+, the aggregate gain over the matched uniform prior is modest (2.1--2.6 percentage points), but the early learning curves show a pronounced advantage: after 50~episodes, O-STaR reaches 65--71\% cumulative success across households, compared with 38--41\% for the uniform prior~(\figref{fig:homer_cumulative_success}).
    The benefit is substantially larger on DynamicTHOR, where semantic initialization improves search success by 23.7--33.0 percentage points across the evaluated noise and difficulty conditions, with larger gains in hard scenes.
    In DynamicTHOR and HOMER+, O-STaR with IO significantly ($p<0.05$) outperformed Uniform Prior + belief + IO across all noise settings for success, visits, and episode time.
    These results indicate that the LLM prior is most valuable before sufficient household-specific evidence has accumulated, after which Uniform prior approaches the success rate of O-STaR.

    \noindent\textbf{RQ2 Target-only belief updates:}
    Across both datasets and all noise settings, target-only belief updates outperform the Static and Push-back LLM rankings, with gains of 16.48--20.39~\mbox{\% points} over Static LLM on HOMER+ and smaller but consistent gains on DynamicTHOR.
    Starting from the same semantic ranking, Static LLM saturates or even deteriorates, and Push-back LLM improves only gradually, whereas O-STaR continues to improve as evidence accumulates~(\figref{fig:homer_cumulative_success}).
    O-STaR without IO significantly ($p<0.05$) outperformed Static LLM across all DynamicTHOR conditions and metrics, except for success in Normal–FN5 and search time in Hard–FN5+FP5, and outperformed ($p<0.05$) both Static and Push-back LLM across all HOMER+ noise settings and metrics.
    These results show that persistent belief updates adapt substantially better to household-specific object searches than rank-based LLM methods.

    \noindent\textbf{RQ3 Incidental observations:}
    Incidental observations provide a further consistent gain over target-only belief updates, improving search success by 4.8--7.4 percentage points on HOMER+ and 3.5--6.5 percentage points on DynamicTHOR across all evaluated noise and difficulty conditions~(\tabref{tab:homer_results}).
    This shows that observations acquired during previous searches provide useful cross-query evidence.

    \subsection{Real-World Experiments}
    \label{sec:exp_robot}
    We evaluate O-STaR on a Toyota HSR-C robot~\cite{yamamoto2018human} in a laboratory mock-up household environment to assess geometric reasoning and real-world object search.
    The experiments evaluate whether LLM-derived size priors enable conservative pruning of physically infeasible compartments and whether affordance-conditioned viewpoints support reliable inspection of exposed surfaces (\texttt{ON\_SURFACE})  and open interiors (\texttt{INSIDE\_OPEN}).
    We first construct an offline 3D scan following the procedure adapted from Spot-Compose~\cite{lemke2023spotcompose} and extract furniture-anchor bounding boxes using Mask3D~\cite{schult2023mask3d}.
    The resulting scene graph is then annotated with inspection affordances using DeepSeek LLM query~\cite{xu2026deepseek}.
    SAM3~\cite{carion2025sam3}, combined with registered depth, is then used to segment concealed compartments and associated manipulation elements, including doors, drawers, knobs, and handles, which are registered in the furniture-anchored scene graph.
    We use the HSR-C's onboard Xtion PRO LIVE \mbox{RGB-D} sensor for perception, with SAM3 serving as the open-vocabulary detector during object-search trials.
    O-STaR runs on Ubuntu~22.04 with \mbox{ROS~2 Humble} on an Intel \mbox{Core i9-14900K}, \mbox{RTX~4080} GPU, and 64~GB RAM; the GPU is required only for SAM3 inference, while the remaining \mbox{O-STaR} components can execute on CPU-only hardware.

    \noindent\textbf{RQ4 Geometric Feasibility Check:}
    We first evaluate size-prior robustness on 10 YCB~\cite{calli2017yale} objects, demonstrating agreement between LLM-derived and measured-size fit ratings for 106 of 110 object--compartment pairs (96.4\%).
    Because YCB dimensions are public, we additionally evaluate 11 randomly selected user-supplied household objects with potentially variable dimensions.
    Across the 11 object queries with reference fit constraints, hard pruning yields 14 true positives, \textbf{0 false positives}, and 38 false negatives.
    Of these false negatives, 27 (71.1\%) are still rated \emph{low}, such that 41 of 52 infeasible pairs (78.8\%) are either pruned or down-weighted.
    Importantly, no feasible location is hard-pruned, indicating conservative geometric filtering while prioritizing feasible concealed spaces over infeasible ones.

    \noindent\textbf{RQ5 Real-world Object Search using Inspection Affordances:}
    \begin{figure}[t]
    \includegraphics[width=0.99\columnwidth]{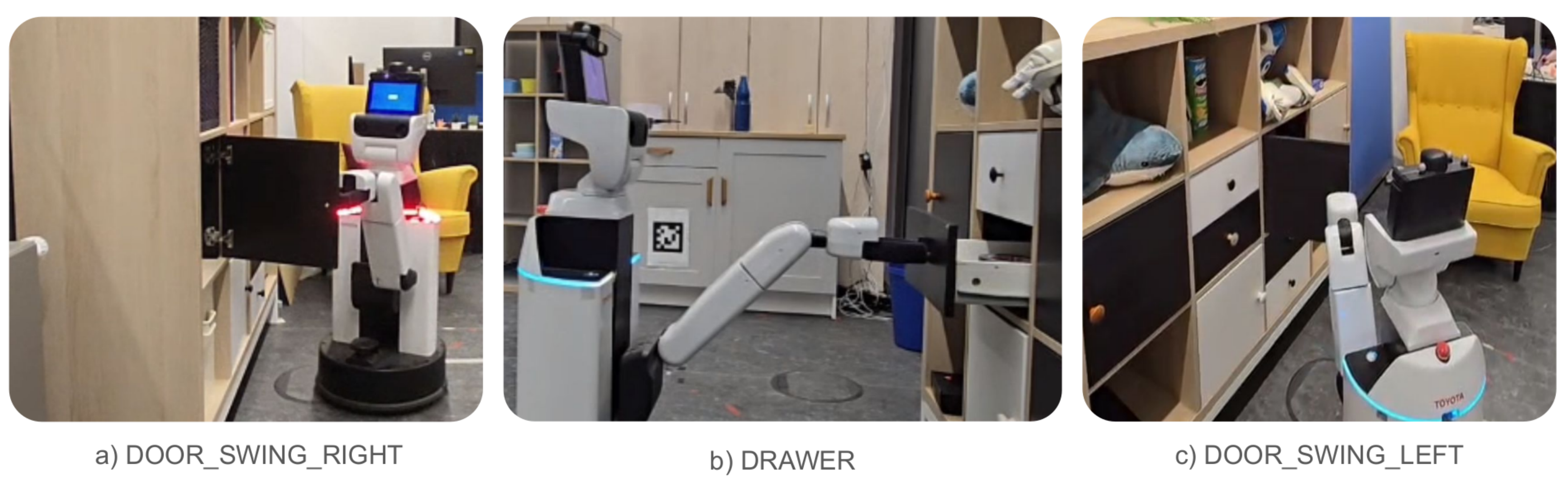}
    \caption{\small{Real-world inspection of concealed search locations.
      \mbox{O-STaR} uses the inferred manipulation affordance to access door-enclosed compartments with different opening directions and drawers before inspecting their interiors for the target object.}}
          \label{fig:concealed_search}
      \vspace{-16px}
    \end{figure}
    \begin{table}[t]
    \raggedright
    \scriptsize
    \setlength{\tabcolsep}{3.5pt}
    \renewcommand{\arraystretch}{1.1}
    \begin{tabular}{@{}lccc@{}}
        \toprule
        \textbf{Method} &
        \textbf{Success} &
        \shortstack{\textbf{Visits}} &
        \shortstack{\textbf{Time}\textbf{(s)}}\\
        \midrule
        Centered &
        8/15 (53.3\%) &
        3.20{$\pm$}2.21 &
        \textbf{110.94}{$\pm$}80.96\\
        \midrule
        Nominal Affordance &
        11/15 (73.3\%) &
        2.93{$\pm$}2.19 &
        191.92{$\pm$}176.34\\
        \midrule
        \shortstack[l]{Affordance-conditioned Clutter-aware} &
        \textbf{13/15 (86.7\%)} &
        \textbf{2.20}{$\pm$}1.57&
         142.16{$\pm$}139.19\\
        \bottomrule
    \end{tabular}
        \caption{\small {Real-robot search results over 15 trials per method. Visits and time are reported per search episode as mean $\pm$ standard deviation.}}
    \label{tab:real_robot_results}
    \vspace{-20px}
    \end{table}
    Following the personality-conditioned household simulation framework of PerSim~\cite{li2026personalize}, we generate a deterministic sequence of 30 temporal scene-graph states over six directly inspectable locations comprising four exposed surfaces and two open shelves.
    Using PerSim-derived placements reduces experimenter bias from manually designed object--location sequences.
    From the 11 household objects used in RQ4, we select five objects---a Pringles can, joystick controller, football, purple binder, and handbag---and search for each three times. 
    Before each trial, all five objects are staged according to the corresponding scene-graph state, while the robot searches only for the scheduled target.
    We compare three inspection strategies: a furniture-centered viewpoint, a nominal affordance-conditioned viewpoint, and affordance-conditioned clutter-aware viewpoints under identical conditions, with independent persistent belief states initialized from the same prior.

    As shown in~\tabref{tab:real_robot_results}, the furniture-centered baseline succeeds in 8/15 trials~(53.3\%), while scene-graph affordances enable nominal affordance-conditioned inspection, increasing success to 11/15~(73.3\%).
    Affordance-conditioned clutter-aware refinement further increases success to 13/15 searches~(86.7\%).
    It also requires the fewest visits, with 2.20 visits compared with 3.20 and 2.93, respectively.
    These real-world experiments demonstrate that O-STaR transfers its sparse scene-graph representation and search reasoning to physical robotic object search, while affordance-conditioned inspection and belief-aware recovery improve search success over naive furniture-centered views.
    Furthermore, as shown in~\figref{fig:concealed_search}, we performed qualitative real-world searches in concealed spaces with geometric feasibility checks and execution of perception-inferred manipulation affordances for interior inspection.

    \section{Conclusion}
    \label{sec:conclusion}
    We presented O-STaR, an open-vocabulary object-search framework for persistent belief-driven reasoning over sparse, furniture-anchored scene graphs.
    O-STaR combines language-model semantic initialization with Dirichlet--categorical updates from target and incidental observations.
    Long-term experiments show that semantic priors improve early search, while persistent updates outperform static and push-back LLM rankings as evidence accumulates.
    The uniform-prior ablation further shows that, given sufficient observations, the belief formulation can learn effective object--location patterns without semantic initialization.
    Incidental observations provide additional gains without dedicated inspections.
    In real-world experiments, LLM-derived size priors conservatively prune infeasible locations, while scene-graph affordances support embodied object search through affordance-conditioned inspection and clutter-aware refinement.
    Overall, O-STaR enables robot-agnostic object search directly on a sparse scene graph without requiring a dense 3D semantic map.

    \section{Acknowledgments}
    ChatGPT~5.6 Sol was used for language editing, grammatical checks, and graphical refinement of~\figref{fig:motivation}.
    \bibliographystyle{IEEEtran}
    \balance
    \bibliography{references}

@STRING{ai                      = {Artificial Intelligence}}

@STRING{arxiv 	                = {arXiv preprint}}

@STRING{corl                    = {Proc.~of Conf.~on Robot Learning (CoRL)}}

@STRING{eccv                    = {Proc.~of the Europ.~Conf.~on Computer Vision (ECCV)}}

@STRING{humanoids               = {Proc.~of the IEEE-RAS Intl.~Conf.~on Humanoid Robots}}

@STRING{icra                    = {Proc.~of the IEEE Intl.~Conf.~on Robotics \& Automation (ICRA)}}

@STRING{ijcai                   = {Proc.~of the Intl.~Conf.~on Artificial Intelligence (IJCAI)}}

@STRING{ijrr                    = {Intl.~Journal~of Robotics Research (IJRR)}}

@STRING{iros                    = {Proc.~of the IEEE/RSJ Intl.~Conf.~on Intelligent Robots and Systems (IROS)}}

@STRING{ral                     = {IEEE Robotics and Automation Letters (RA-L)}}

@STRING{ram                     = {IEEE Robotics and Automation Magazine (RAM)}}

@STRING{ras                     = {Journal on Robotics and Autonomous Systems (RAS)}}

@STRING{rss                     = {Proc.~of Robotics: Science and Systems (RSS)}}

@STRING{tro                     = {IEEE Trans.~on Robotics (TRO)}}

@phdthesis{miller2018understanding,
  title={Understanding clutter: Geographies of everyday homes and objects},
  author={Miller, Alexander James},
  year={2018},
  school={University of Leeds}
}

@inproceedings{gu2024conceptgraphs,
  title={Conceptgraphs: Open-vocabulary 3d scene graphs for perception and planning},
  author={Gu, Qiao and Kuwajerwala, Ali and Morin, Sacha and Jatavallabhula, Krishna Murthy and Sen, Bipasha and Agarwal, Aditya and Rivera, Corban and Paul, William and Ellis, Kirsty and Chellappa, Rama and others},
  booktitle=icra,
  year={2024}
}

@article{lemke2023spotcompose,
  title={Spot-compose: A framework for open-vocabulary object retrieval and drawer manipulation in point clouds},
  author={Lemke, Oliver and Bauer, Zuria and Zurbr{\"u}gg, Ren{\'e} and Pollefeys, Marc and Engelmann, Francis and Blum, Hermann},
  journal=arxiv,
  year={2024}
}

@article{liu2024dynamem,
  title={DynaMem: Online Dynamic Spatio-Semantic Memory for Open World Mobile Manipulation},
  author={Liu, Peiqi and Guo, Zhanqiu and Warke, Mohit and Chintala, Soumith and Paxton, Chris and Shafiullah, Nur Muhammad Mahi and Pinto, Lerrel},
  journal=arxiv,
  year={2024}
}

@article{liu2024okrobot,
  title={OK-Robot: What Really Matters in Integrating Open-Knowledge Models for Robotics},
  author={Liu, Peiqi and Orru, Yaswanth and Vakil, Jay and Paxton, Chris and Shafiullah, Nur Muhammad Mahi and Pinto, Lerrel},
  journal=arxiv,
  year={2024}
}

@article{qiu2024jsr1,
  title={Open-vocabulary Mobile Manipulation in Unseen Dynamic Environments with 3D Semantic Maps},
  author={Qiu, Dicong and Ma, Wenzong and Pan, Zhenfu and Xiong, Hui and Liang, Junwei},
  journal=arxiv,
  year={2024}
}

@article{honerkamp2024momallm,
  title={Language-grounded dynamic scene graphs for interactive object search with mobile manipulation},
  author={Honerkamp, Daniel and B{\"u}chner, Martin and Despinoy, Fabien and Welschehold, Tim and Valada, Abhinav},
  journal=ral,
  year={2024},
  publisher={IEEE}
}

@inproceedings{rozenberszki2022language,
    title={Language-Grounded Indoor 3D Semantic Segmentation in the Wild},
    author={Rozenberszki, David and Litany, Or and Dai, Angela},
    booktitle = eccv,
    year={2022}
}

@inproceedings{schult2023mask3d,
  title={Mask3d: Mask transformer for 3d semantic instance segmentation},
  author={Schult, Jonas and Engelmann, Francis and Hermans, Alexander and Litany, Or and Tang, Siyu and Leibe, Bastian},
  booktitle=icra,
  year={2023},
  organization={IEEE}
}

@inproceedings{zheng2023system,
  title={A System for Generalized 3D Multi-Object Search},
  author={Zheng, Kaiyu and Paul, Anirudha and Tellex, Stefanie},
  booktitle=icra,
  year={2023},
  organization={IEEE}
}

@article{werby23hovsg,
Author = {Abdelrhman Werby and Chenguang Huang and Martin Büchner and Abhinav Valada and Wolfram Burgard},
Title = {Hierarchical Open-Vocabulary 3D Scene Graphs for Language-Grounded Robot Navigation},
Year = {2024},
journal = rss,
}

@book{hochschild2012second,
  title={The second shift: Working families and the revolution at home},
  author={Hochschild, Arlie and Machung, Anne},
  year={2012},
  publisher={Penguin}
}

@inproceedings{aydemir2011search,
  title={Search in the real world: Active visual object search based on spatial relations},
  author={Aydemir, Alper and Sj{\"o}{\"o}, Kristoffer and Folkesson, John and Pronobis, Andrzej and Jensfelt, Patric},
  booktitle=icra,
  year={2011},
  organization={IEEE}
}

@inproceedings{kunze2014using,
  title={Using qualitative spatial relations for indirect object search},
  author={Kunze, Lars and Doreswamy, Keerthi Kumar and Hawes, Nick},
  booktitle=icra,
  year={2014},
  organization={IEEE}
}

@inproceedings{hanheide2011probabilistic,
  author    = {Hanheide, Marc and Gretton, Charles and Dearden, Richard and Hawes, Nick and Wyatt, Jeremy L. and Pronobis, Andrzej and Aydemir, Alper and G{\"o}belbecker, Moritz and Zender, Hendrik},
  title     = {Exploiting Probabilistic Knowledge under Uncertain Sensing for Efficient Robot Behaviour},
  booktitle = ijcai,
  year      = {2011},
}

@inproceedings{wandzel2019multi,
  title={Multi-object search using object-oriented pomdps},
  author={Wandzel, Arthur and Oh, Yoonseon and Fishman, Michael and Kumar, Nishanth and Wong, Lawson LS and Tellex, Stefanie},
  booktitle=icra,
  year={2019},
  organization={IEEE}
}

@inproceedings{zeng2020semantic,
  title={Semantic linking maps for active visual object search},
  author={Zeng, Zhen and R{\"o}fer, Adrian and Jenkins, Odest Chadwicke},
  booktitle={2020 IEEE international conference on robotics and automation (ICRA)},
  year={2020},
  organization={IEEE}
}

@article{hawes2017strands,
  title={The strands project: Long-term autonomy in everyday environments},
  author={Hawes, Nick and Burbridge, Christopher and Jovan, Ferdian and Kunze, Lars and Lacerda, Bruno and Mudrova, Lenka and Young, Jay and Wyatt, Jeremy and Hebesberger, Denise and Kortner, Tobias and others},
  journal=ram,
  volume={24},
  number={3},
  year={2017},
  publisher={IEEE}
}

@article{krajnik2017fremen,
  title={Fremen: Frequency map enhancement for long-term mobile robot autonomy in changing environments},
  author={Krajn{\'\i}k, Tom{\'a}{\v{s}} and Fentanes, Jaime P and Santos, Joao M and Duckett, Tom},
  journal=tro,
  volume={33},
  number={4},
  year={2017},
  publisher={IEEE}
}

@inproceedings{patel2023predicting,
    title={Predicting Routine Object Usage for Proactive Robot Assistance}, author={Maithili Patel and Aswin Gururaj Prakash and Sonia Chernova}, booktitle=corl, 
    year={2023}
}

@inproceedings{patel2022proactive,
  title={Proactive Robot Assistance via Spatio-Temporal Object Modeling},
  author={Patel, Maithili and Chernova, Sonia},
  booktitle=corl,
  year={2023},
  organization={PMLR}
}

@article{carion2025sam3,
      title={SAM 3: Segment Anything with Concepts},
      author={Nicolas Carion and others},
      journal={arXiv preprint arXiv:2511.16719},
      year={2025},
      eprint={2511.16719},
      archivePrefix={arXiv},
      primaryClass={cs.CV},
}

@article{wang2026dynamicthor,
  title={{DynamicTHOR}: A Scalable Dataset of Human-Centric Dynamic Scenes for Embodied AI},
  author={Wang, Chenxu and Wang, Dunzheng and Wang, Dong and Li, Xinghang and Li, Yongchang and Liu, Huaping},
  journal={Scientific Data},
  volume={13},
  number={1},
  year={2026},
}

@article{bogenberger2026glasses,
  title={Where Did I Leave My Glasses? Open-Vocabulary Semantic Exploration in Real-World Semi-Static Environments},
  author={Bogenberger, Benjamin and Harrison, Oliver and Dahanaggamaarachchi, Orrin and Brunke, Lukas and Qian, Jingxing and Zhou, Siqi and Schoellig, Angela P.},
  journal=ral,
  volume={11},
  number={3},
  year={2026}
}

@inproceedings{zheng2021multi,
  title={Multi-Resolution {POMDP} Planning for Multi-Object Search in 3D},
  author={Zheng, Kaiyu and Sung, Yoonchang and Konidaris, George and Tellex, Stefanie},
  booktitle=iros,
  year={2021},
}

@article{xu2026deepseek,
  title={Deepseek-v4: Towards highly efficient million-token context intelligence},
  author={Xu, Anyi and Lin, Bangcai and Xue, Bing and Wang, Bingxuan and Xu, Bingzheng and Wu, Bochao and Zhang, Bowei and Lin, Chaofan and Dong, Chen and Ling, Chenchen and others},
  journal={arXiv preprint arXiv:2606.19348},
  year={2026}
}

@incollection{yamamoto2018human,
  title={Human support robot (HSR)},
  author={Yamamoto, Takashi and Nishino, Tamaki and Kajima, Hideki and Ohta, Mitsunori and Ikeda, Koichi},
  booktitle={ACM SIGGRAPH 2018 emerging technologies},
  year={2018}
}

@article{loo2025openscenegraphs,
  author  = {Loo, Joel and Wu, Zhanxin and Hsu, David},
  title   = {Open Scene Graphs for Open-World Object-Goal Navigation},
  journal = ijrr,
  year    = {2025},
  publisher={SAGE Publications Sage UK: London, England}

}

@inproceedings{chikhalikar2025enhancing,
  title={Enhancing object search in indoor spaces via personalized object-factored ontologies},
  author={Chikhalikar, Akash and Ravankar, Ankit A and Luces, Jose Victorio Salazar and Hirata, Yasuhisa},
  booktitle=iros,
  year={2025},
  organization={IEEE}
}

@article{xie2026osmag,
  title={osmag-llm: Zero-shot open-vocabulary object navigation via semantic maps and large language models reasoning},
  author={Xie, Fujing and Schwertfeger, S{\"o}ren and Blum, Hermann},
  journal=ral,
  year={2026},
  publisher={IEEE}
}

@article{schmalstieg2023learning,
  author  = {Schmalstieg, Fabian and Honerkamp, Daniel and Welschehold, Tim and Valada, Abhinav},
  title   = {Learning Hierarchical Interactive Multi-Object Search for Mobile Manipulation},
  journal = ral,
  year    = {2023},
  publisher={IEEE}
}

@article{koc2024objectaware,
  author  = {Koc, Cagatay and Sariel, Sanem},
  title   = {Object-Aware Interactive Perception for Tabletop Scene Exploration},
  journal = ras,
  year    = {2024},
}

@inproceedings{dengler2025efficient,
  author    = {Dengler, Nils and M{\"u}cke, Jesper and Menon, Rohit and Bennewitz, Maren},
  title     = {Efficient Manipulation-Enhanced Semantic Mapping With Uncertainty-Informed Action Selection},
  booktitle = humanoids,
  year      = {2025},
}

@article{mahdi2026relational,
  title={Relational semantic reasoning on 3d scene graphs for open world interactive object search},
  author={Mahdi, Imen and Cassinelli, Matteo and Despinoy, Fabien and Welschehold, Tim and Valada, Abhinav},
  journal={arXiv preprint arXiv:2603.05642},
  year={2026}
}

@article{li2026personalize,
  title={When to Personalize Household Object Search: A Rigidity-Gated Hybrid Policy},
  author={Li, Xianyao and Wang, Yuhai and Xiao, Hu and Smith, Kaleb and Ye, Gilbert Yang and Du, Eric Jing},
  journal={arXiv preprint arXiv:2607.00022},
  year={2026}
}

@article{calli2017yale,
  title={Yale-CMU-Berkeley dataset for robotic manipulation research},
  author={Calli, Berk and Singh, Arjun and Bruce, James and Walsman, Aaron and Konolige, Kurt and Srinivasa, Siddhartha and Abbeel, Pieter and Dollar, Aaron M},
  journal=ijrr,
  volume={36},
  number={3},
  year={2017},
  publisher={SAGE Publications Sage UK: London, England}
}

@inproceedings{menon2025open,
  title={Open-Vocabulary and Semantic-Aware Reasoning for Search and Retrieval of Objects in Dynamic and Concealed Spaces},
  author={Menon, Rohit and Schmiede, Yasmin and Bennewitz, Maren and Blum, Hermann},
  booktitle={Perception and Planning for Mobile Manipulation in Changing Environments (IROS 2025) Workshop},
  year={2025}
}
    \end{document}